\documentclass[11pt]{article}

\usepackage{EMNLP2022}

\usepackage{times}
\usepackage{latexsym}
\usepackage{graphicx}

\usepackage[T1]{fontenc}

\usepackage[utf8]{inputenc}

\usepackage{microtype}

\usepackage{inconsolata}

\usepackage{hyperref} 
\usepackage{booktabs} 
\usepackage{enumitem}
\setitemize{noitemsep,topsep=0pt,parsep=0pt,partopsep=0pt}

\definecolor{lightgray}{gray}{0.94}

\title{Dungeons and Dragons as a Dialog Challenge for Artificial Intelligence}

\author{Chris Callison-Burch\thanks{*Denotes equal contribution} \\
University of Pennsylvania \\
\texttt{ccb@upenn.edu} \\\And
  Gaurav Singh Tomar$^*$ \\
  Google Research\\
  \texttt{gtomar@google.com} \\\And
  Lara J. Martin\\
  University of Pennsylvania \\\AND
  Daphne Ippolito\\
  University of Pennsylvania \\ Google Research\\\And
  Suma Bailis\\
  Google Research\\\And
  David Reitter\\
  Google Research}

\newcommand\DnD{D\&D~}

\newcommand\DnDnospace{D\&D}

\begin{document}
\maketitle

\begin{abstract}
AI researchers have posited Dungeons and Dragons (D\&D) as a challenge problem to test systems on various language-related capabilities. In this paper, we frame D\&D specifically as a dialogue system challenge, where the tasks are to both generate the next conversational turn in the game and predict the state of the game given the dialogue history.  We create a gameplay dataset consisting of nearly 900 games, with a total of 7,000 players, 800,000 dialogue turns, 500,000 dice rolls, and 58 million words.  We automatically annotate the data with partial state information about the game play.  We train a large language model (LM) to generate the next game turn, conditioning it on different information.  The LM can respond as a particular character or as the player who runs the game—i.e., the Dungeon Master (DM). It is trained to produce dialogue that is either in-character (roleplaying in the fictional world) or out-of-character (discussing rules or strategy).  We perform a human evaluation to determine what factors make the generated output plausible and interesting.  We further perform an automatic evaluation to determine how well the model can predict the game state given the history and examine how well tracking the game state improves its ability to produce plausible conversational output.  

\end{abstract}

\section{Introduction}

Artificial Intelligence has a long and rich history of using games as challenge problems that lead to advances in the field. In many cases, AI game-playing systems have gone on to rival human champions of the game.
%
%
%
%
Dungeons and Dragons has been identified as an appropriate challenge for the next stage of artificial intelligence \cite{Ellis2017, Louis2018, Martin2018INT}.
\newcite{Ellis2017} proposed open-ended creative games like \DnD~as the next challenge for AI after the human-level successes of AI at Chess and Go, which are zero-sum, deterministic, sequential two-player games with perfect information. \newcite{Louis2018} understood the importance of narrative in natural language processing (NLP) and generation (NLG). In particular, they saw how cooperative story generation between humans already exists in these games and can be used for automated generation. \newcite{Martin2018INT} outlined some of the specific challenges \DnD~presents to the NLP community; such as a state of the game world distributed across the Dungeon Master (DM) and other players or dealing with the intrinsic rewards players get from taking certain actions that would not necessarily provide them with points in the game.

\DnD involves multiple players who roleplay characters in a fantasy setting, guided by a Dungeon Master who sets obstacles and adventures and plays as monsters. 
In roleplaying games like Dungeons and Dragons, the gameplay happens through language rather than moves on a game board, making it an interesting domain for NLP research.
To have an AI successfully play  \DnD, it would require abilities like

\begin{itemize}
    \item Language generation (multi-party dialog, generating descriptions of the world/actions, storytelling)
    \item Language understanding (knowledge acquisition and representation, state tracking, automated reasoning)
    \item Planning / strategic play during battles (similar to chess or go)
\end{itemize}
Appendix \ref{example-game-play} gives an example of \DnD gameplay and the AI challenges presented by it. 

Is it possible to design an AI system that is capable of playing a game of \DnD either as a character in the game or as the Dungeon Master using current AI technology?  We argue that now is the perfect time for this challenge, since large scale neural language models like GPT have shown impressive generation results \citep{brown2020language}, and since incorporating neural LMs into a game setting both exercises their strengths and exposes their weaknesses. 

In this paper, we introduce a new dataset of “actual play” game transcripts. Each turn is labeled with game state variables like character information and whether the conversational turn was in-character or out-of-character.   Our data is a novel, large scale, real-world conversational dataset.  It is unique in that the dialog turns are generated entirely through player collaboration and written interaction in a multi-player game. We propose our dataset as a challenge for dialogue systems for the following reasons: 
\begin{itemize}
    \item It is naturally occurring conversational dialog that covers a spectrum of task oriented and non-task oriented (e.g. chit chat) dialog.
    \item It is strongly history dependent – a substantive criticism of recent dialog datasets is their history independence \citep{DBLP:journals/corr/abs-2004-10473}.
    \item It has many participants in the conversation, since there are several players in the game.
    \item It conveys narrative elements including descriptions of events that denote changes in the state of the game.
\end{itemize}
Unlike existing dialog datasets, our data reflects the challenging nature of the \DnD game as a multi-party dialogue with creative roleplaying and underlying game states. 

\section{Tasks}

We trained a large language model (LLM) to perform two tasks: {\bf Next Utterance Prediction} and {\bf Game State Tracking}.  

\paragraph{Next Utterance Prediction.} We trained our language model on a corpus of human conversations (see Section \ref{sec:dataset}) to predict the next utterance.   We varied the conditioning information to examine the effects on the quality of predicted next utterance. In all variations, we included the conversational history as input.  
Given the conversational input (and other input in the variant models), the LLM must generate the next utterance, such that it is both interesting and a plausible next turn in the \DnD game.

\paragraph{Game State Tracking.} In this task, rather than producing the next utterance, we had the model predict the game state for a given dialogue turn in the conversation. We have kept the state definition similar to task-oriented dialogue state tracking (DST).  In DST, the dialogue state is a collection of slot-value pairs. In our case, each slot is a state variable feature related to \DnD games. Our target slot values do not need to appear as a word in the dialogue context. We track several game states aspects including some that remain relatively static throughout the game (character attributes like their pronouns, class, fantasy race, and their inventory), some that change periodically (like being in combat or out of combat), and some that change from turn to turn (like what action the player is taking).

\section{Dataset}
\label{sec:dataset}
For this paper, we have created a novel dataset for our dialogue-oriented test of AI's ability to play Dungeons \& Dragons. We scraped
Play-By-Post data from a web forum\footnote{\url{https://www.dndbeyond.com/forums/d-d-beyond-general/play-by-post}}   where people play by taking turns posting on the forum to describe their move.  

\begin{figure*}
    \centering
    \includegraphics[width=\linewidth]{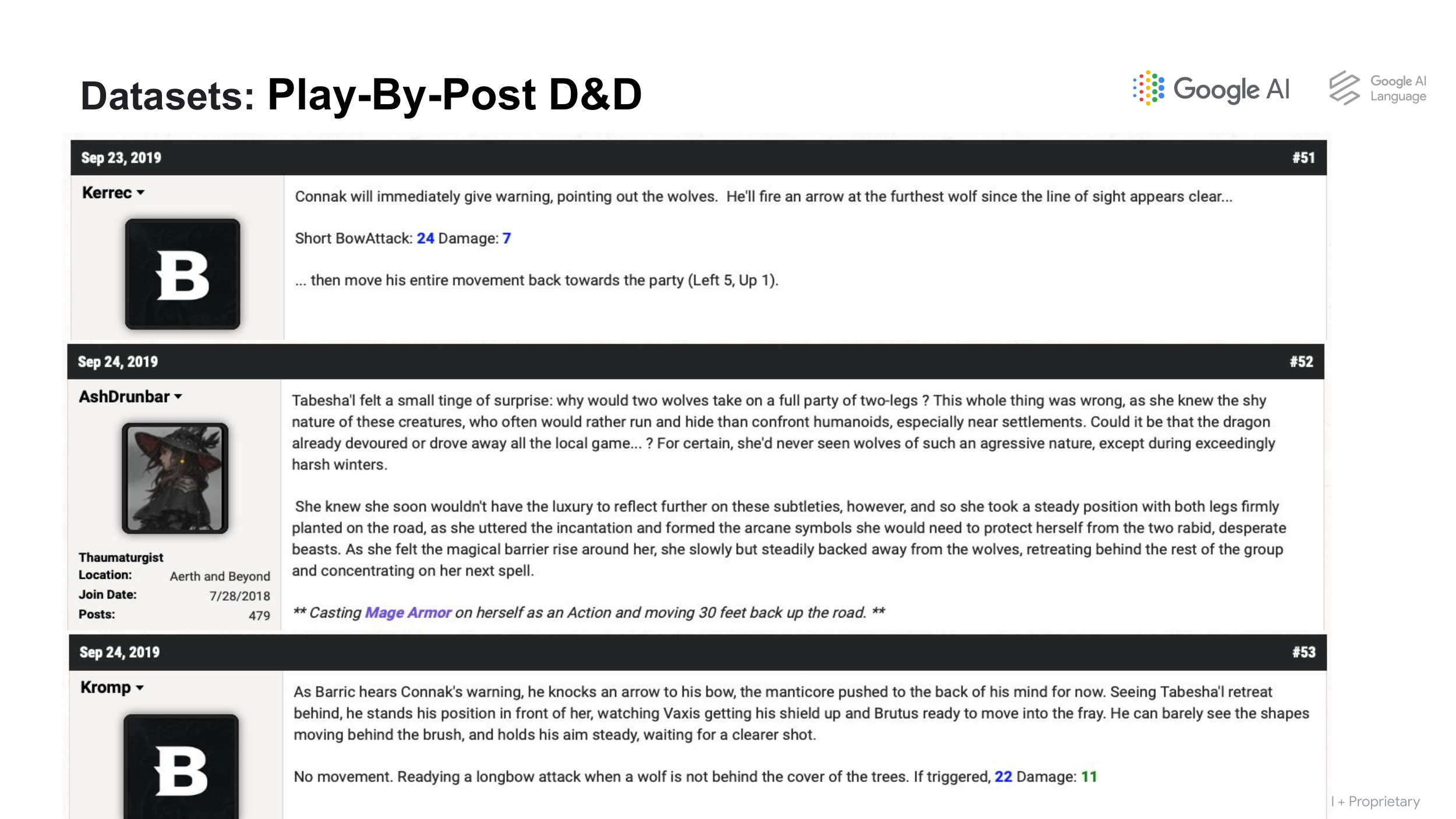}
    \caption{Example of 3 turns in the \DnD Beyond play-by-post forum}
    \label{fig:play-by-post-example}
\end{figure*}

Figure \ref{fig:play-by-post-example} shows an example of part of the gameplay from the play-by-post  forums from \DnD Beyond.
\DnD Beyond provides a mechanism in its forum to roll dice using a “roll” tag.  Their dice roller allows players to conduct the rolls that are used for \DnD ability checks and in combat.  

Table \ref{tab:play-by-post-stats} summarizes the amount of play-by-post data that we collected from the \DnD Beyond website (with permission from the company).

\begin{table}
\centering
\begin{tabular}{lr}
\hline
\multicolumn{2}{c}{{\bf Play-By-Post Corpus }}\\
\hline
Number of campaigns	&	896	\\
Average players per campaign	&	8	\\
Average turns per campaign	&	910	\\
Average words per campaign	&	64,941	\\
Total turns	&	815,106	\\
Total words	&	58,187,526	\\
\hline
Average dice rolls per campaign	&	594	\\
Total dice rolls	&	532,270	\\
\hline
\end{tabular}
\caption{Statistics for our play-by-post corpus}\label{tab:play-by-post-stats}
\end{table}

\subsection{Heuristic annotation of game states}\label{heuristic-annotation}

We designed a set of rule-based heuristics to extract game state information from the play by post.  These were implemented using regular expressions and NLP tools like named entity recognizers \cite{allennlp}.  Although this heuristically extracted information is not perfect, it provides a reasonable approximation of the game state.  It is useful for testing whether large language models can benefit from inclusion of complex state information for next utterance prediction and whether LLMs can be used for state tracking.  We designed rules to extract state information relating to character properties, combat and player actions.

\begin{table*}
\small
\begin{tabular}{p{.13\linewidth}p{.36\linewidth}p{.42\linewidth}}
\toprule
Control Feature	&	Description	&	Expected Impact on Model's Output	\\ \hline
Player ID	&	Player writing a given dialog turn	&	Connects the current turn to the player's previous turns, which is important in multi-party conversations.	\\
IC versus OOC	&	Whether a player is in-character or out-of-character for a given dialog turn	&	Changes whether the generated text is more like descriptive text found in a novel, or more like a discussion of rules and strategies.	\\
Character Name	&	Name of the character being played by the player of a given dialog turn	&	IC descriptions use the character's name.	\\
Character Class	&	 \DnD classes	&	Character classes perform different actions (e.g. wizards cast spells, thieves pick locks)	\\
Character Race	&	 \DnD fantasy races	&	Different physical characteristics (e.g. halflings are small, dragonborn have scales).	\\
Character Pronouns	&	The character's pronouns	&	Uses the correct pronouns when describing the character.	\\
Character Actions	&	List of actions taken by the character in the current turn	&	Allows a description to be generated for a given action.  The action can be thought of as a goal for the description.	\\
Combat	&	Whether the players are currently engaged in combat or not during a given dialog turn	&	Affects the likelihood of actions (e.g. attacks are more likely during combat and investigations checks are more likely outside of combat)	\\
\bottomrule
\end{tabular}
\normalsize
\caption{Our LLMs are conditioned on a variety of control features that allow the models to better learn what kind of text to generate for the next utterance prediction task}\label{tab:control-features}
\end{table*}

\paragraph{Character properties}
\begin{itemize}
    \item Name: Perform NER on all the player's turns in a campaign.  The character's name is assigned to be the player's most frequently mentioned name, on the assumption that they tend to describe their own character's actions.
    \item Class: Count how many times each \DnD class\footnote{\url{https://www.dndbeyond.com/classes}} 
    is mentioned by each player.  Most frequently mentioned class is their character's class.
    \item Race: On a player's first turn, check whether any of the \DnD fantasy races\footnote{\url{https://www.dndbeyond.com/races}} 
    are mentioned.  Assign it to character.  If not, guess based on the most frequently mentioned race. 
    \item Pronouns: Count pronouns mentioned by a player.  Assign their character's pronouns to be the most frequent pronouns used by the player.
    \item Inventory: Use regex to match items occurring after character's personal pronouns (e.g. her sword).
    \item Spells known: Regex that matches cast followed by a spell name
\end{itemize}
The DM is assumed to be the player who has the first post in the game.  The DM's entries in the dataset are scrubbed of other character properties, since they play multiple NPCs (non-player characters) and monsters.

\paragraph{Combat}
\begin{itemize}
    \item We detect the start of combat when there is a roll for initiative, or when there are attack rolls before initiative (from surprise attacks).
    \item Combat continues while there are attack rolls happening.
    \item Combat concludes after there are no rolls for a number of turns.
    \item In a combat span, we extract a list of monsters mentioned, and heuristically guess the number of each kind of monster.
\end{itemize}

\paragraph{Actions}
\begin{itemize}
    \item Dice rolls are marked in \DnD Beyond posts.  We detect the associated actions based on the kind of die used (D20 = a check, other dice are used for calculating damage if an attack check is successful)
    \item We use a regex to match the nearest pattern, which includes attack or a list of abilities like {\it acrobatics, animal handling, arcana, athletics,} etc.
    \item Damage rolls are matched with {\it damage, dmg, cure, heal, healing, points}.
\end{itemize}

Our heuristics resulted in features for around 60\% of all conversational turns. We train a convolutional neural network classifier using these conversational turns to predict all of the above control features for each conversational turn in training data.  Appendix \ref{cnn-state-values} estimates the accuracy of the model's prediction on these state features. 

\subsection{In-Character Versus Out-Of-Character Text}


In addition to labeling the game states in our Play by Post data, we also labeled the text of each turn as being either spoken in-character (IC) or out-of-character (OOC).  To do so, we crawled another Play by Post forum hosted at Giant in the Playground\footnote{\url{https://forums.giantitp.com/forumdisplay.php?3-Play-by-Post-Games}}, 
where play happens on two discussion boards – one in-character and one out-of-characters.  For example, here is an IC post:
\begin{quote}
Kuros pulls the feathered shaft of the arrow back to his cheek winning easily against the resistance of
the bowstring. He pulls a lungful of air to keep himself steady, takes aim at the Bandit with the deer,
and lets fly.
\end{quote}
And here is its corresponding OOC post:
\begin{quote}
Surprise round so only 1 standard or move action. 
Shoot the bow: (1d20+6)[20] vs Flat Footed AC at Bandit 1. 
Damage: (1d8+2)[10]
\end{quote}
We train a classifier to predict IC versus OOC text, and then apply it to each paragraph in our \DnD Beyond forum data.

\section{Models}

For our large language model, we use a 64B parameter version of Google's LaMDA language model \cite{lamda:2022}, which was trained on conversations.  LaMDA is similar to other Transformer-based pre-trained language models like GPT-3.  As with other pre-trained language models \cite{howard-ruder-2018-universal}, LaMDA can be finetuned to different tasks. The two tasks that we finetune LaMDA to perform are game state tracking and response generation.  In both cases, the LLM can be thought of as a function that maps inputs onto an output.  For instance, game state tracking is a language understanding task where the function takes in inputs like $f($current utterance, previous state, history$) \rightarrow$ new state, and response generation is a language generation task where $f($current state, history$) \rightarrow$ next 
utterance.  The LLM functions are trained via the fine tuning process.  

In our experiments we try a variety of different inputs to our LLM functions to see how they enable better learning of the tasks.  
We train our LLMs on the conversation history (which is typical in  dialog modeling) and we also augment the conversations by conditioning other explicit signals.  These conditioning signals can be thought of as sophisticated “control features”, inspired by the CTRL language model \cite{Keskar2019}.  During training, the model learns a relationship between the control features and appropriate responses. In turn, during inference, one can explicitly influence dimensions of the conversation – enabling more compelling dialogue – by setting the values of control features. These control features can be set dynamically, without necessitating finetuning or additional post-processing. Table \ref{tab:control-features} describes the control features we have proposed and describes how they could steer generation.   Note that we use the terms `control features' and `state variables' interchangeably when referring to our next utterance prediction models.

\subsection{Baseline Pre-Training Data}

LaMDA is trained on turn-based conversational data. For a conversation of length $n$, LaMDA takes as input the first $n-1$ turns, and the $n$th turn as the target. 
For all models, we used the 7 most recent conversational turns as input, and predict turn 8.

\subsection{\DnD FineTuning Data}

Here is an example of the data used in our versions of LaMDA that are finetuned to on our  \DnD data.
\newline\newline
\noindent
\small
\begin{tabular}{lp{.70\linewidth}}
{\bf TURN 1}:&\\
Text	&	You attack. You launch some fire onto the goblin closest to the wagon. And with that, he looks like he is on death's door. And the other goblin that you can see, the one that's not in the brush somewhere, just sort of stops in his tracks. What do you do next?	\\
Player ID	&	0	\\
Character	&	Dungeon Master	\\
Race	&	N/A	\\
Class	&	Dungeon Master	\\
Pronouns	&	N/A	\\
Inventory	&	N/A	\\
In combat?	&	Yes	\\
In character?	&	Yes	\\
Action	&	Attack	\\
\end{tabular}
\normalsize

\noindent
\small
\begin{tabular}{lp{.75\linewidth}}
{\bf TURN 2}:&\\
Text	&	I grab my axe and bring it down on the wounded goblin.	\\
Player ID	&	1	\\
Character	&	Magnus	\\
Race	&	Human	\\
Class	&	Fighter	\\
Pronouns	&	he/him	\\
Inventory	&	Axe	\\
In combat?	&	Yes	\\
In character?	&	Yes	\\
Action	&	Attack	\\
\end{tabular}
\normalsize

\subsection{Next Utterance Prediction Models}

\paragraph{LLM-Dialog:}
We call our baseline model LLM-Dialog.  It is a LaMDA dialogue model that does not use not use any \DnD data.

\paragraph{LLM-DND:}

LLM-Dialog that has been finetuned on Play-by-post \DnD gameplay dataset using {\bf no} control features

%

\paragraph{LLM-DND-ALL-CTRL:}

LLM-Dialog that has been finetuned on Play-by-post \DnD gameplay dataset using control features (state variables) for {\bf all} dialog turns including the state variables for {\bf the current turn} the utterance is being predicted for.

%

\paragraph{LLM-DND-PREV-CTRL:}

LLM-Dialog that has been finetuned on Play-by-post \DnD gameplay dataset using control features for all {\bf previous} dialog turns, not including the current turn.


\paragraph{LLM-DND-CURRENT-CTRL:}

LLM-Dialog that has been finetuned on Play-by-post \DnD gameplay dataset using control features (state variables) for  {\bf only the current turn} the utterance is being predicted for.


\subsection{Dev Set Perplexity During Training}
\begin{figure}
    \centering
    \includegraphics[width=\linewidth]{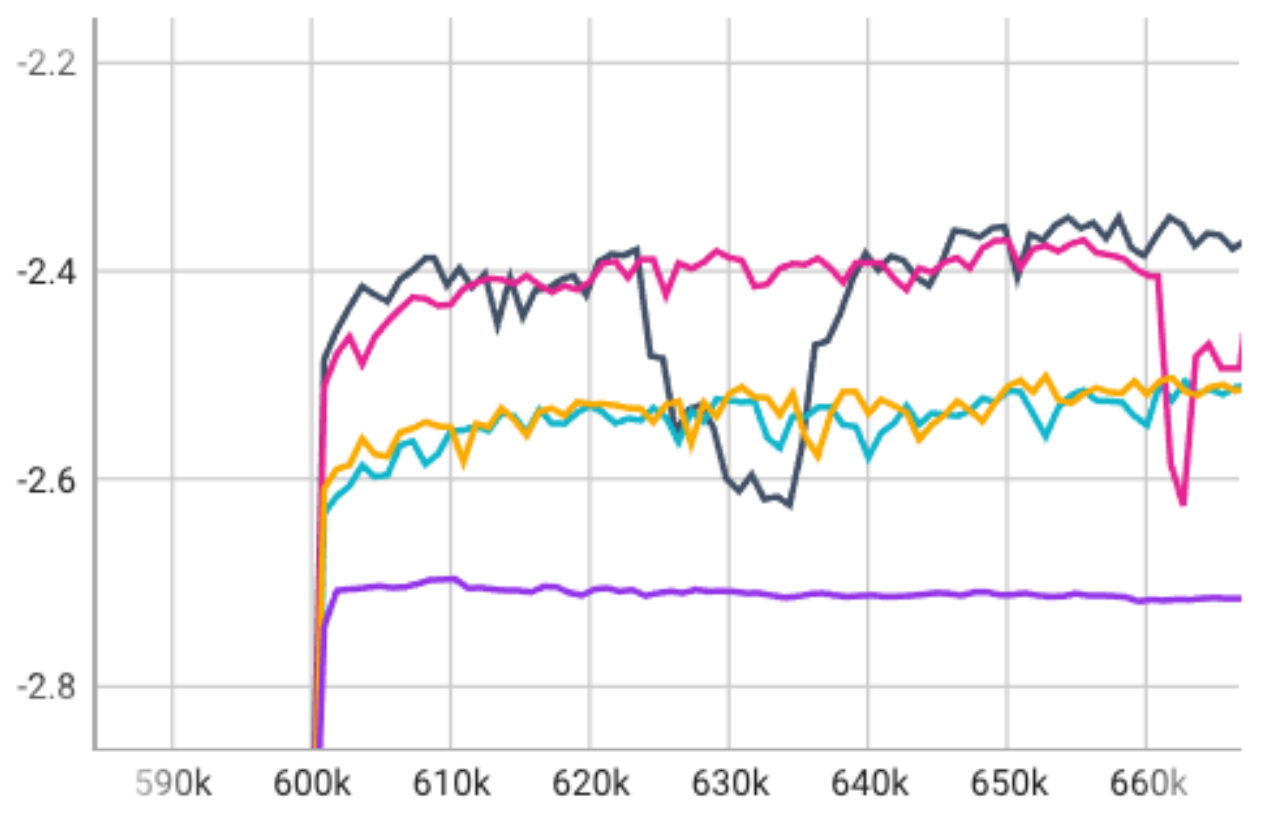}
    \caption{Negative Log Perplexity of our models after pretraining on generic dialogue data for 600k steps, and then finetuning to our data for a further 60k steps.  Colors correspond to the models in Table \ref{token-accuracy}. }\label{fig:perpelxity}
\end{figure}

Each of our models starts from a pretrained LaMDA model trained for 600K steps and then is finetuned for a further 60K steps.   Figure \ref{fig:perpelxity} plots the Negative log perplexity on our development set, and Table \ref{token-accuracy} shows the final perpexity and token accuracies on the dev set.  At the end of finetuning, the models with the best perplexity scores and the best token accuracy scores were  LLM-DND-CURRENT-CTRL and LLM-DND-ALL-CTRL, which used our control features.  

\begin{table}
    \centering
\small
    \begin{tabular}{lcc}
{\bf Model}	&	{\bf Perplexity}	&	{\bf Token Accuracy}	\\\hline
{\color{violet}LLM-Dialog}	&	2.65 & 44.61		\\
{\color{orange}LLM-DND}	&	2.50 & 46.92   \\
{\color{teal}LLM-DND-PREV-CTRL}	&	2.51 & 46.84	\\
{\color{black}LLM-DND-CURR-CTRL}	&	{\bf 2.34} & {\bf 49.67}	\\
{\color{magenta}LLM-DND-ALL-CTRL}	&	2.37 & 49.02	\\
    \end{tabular}
\normalsize
    \caption{Perplexity and token accuracy of our models after finetuning to our data.}
    \label{token-accuracy}
\end{table}

\section{Manual Evaluation}

To evaluate the quality of our models for the task of next utterance prediction in \DnD, we perform a human evaluation. We recruited professional raters to perform a manual evaluation.  They read a version of the content that was provided to the models – the seven turns of conversational history plus a list of players and the names/classes of the characters that they played.  Then they were shown several model outputs for the context (or the “gold”, which was the actual next turn in the game),  The annotators asked to rate each output along the three dimensions, following the evaluation procedure used for the Meena LM \cite{meena}:
\begin{itemize}
    \item Does the response make {\bf sense}? (yes/no)
    \item Is the response {\bf specific}? (yes/no)
    \item How {\bf interesting} is the response? (10 point scale)
\end{itemize}
The full annotator instructions and the annotation interface are given in Appendix \ref{annotation-details}. 

\subsection{Raters}
Because of the specialized nature of the \DnD domain, we recruited 6 professional raters rather than crowd workers to perform the task.  The raters were selected based on their professed interest in the fantasy genre, and on their background with \DnDnospace.  All raters were fantasy fans, and 5 of the 6 had played \DnDnospace.  3 raters had been the DM in a game before. 

\subsection{Inter-Rater Agreement}
Our raters annotated 500 system outputs with 3-way redundancy on each output.  For the binary sense and specific scores, pairwise annotator agreement was 0.8, with a chance-adjusted Randolph Kappa score of 0.6.  For the scalar interestingness scores, the Kendall's Tau correlation was 0.46.

\begin{table}
    \centering
\small
    \begin{tabular}{lccc}
{\bf Model}	&	{\bf Sense}	&	{\bf Specific}	&	{\bf Interest}	\\\hline
{\color{violet}LLM-Dialog}	&	0.81	&	0.85	&	3.57		\\
{\color{orange}LLM-DND}	&	{\bf 0.9}	&	{\bf 0.9}	&	3.91    \\
{\color{teal}LLM-DND-PREV-CTRL}		&	0.86	&	0.88	&	{\bf 3.96}	\\
{\color{black}LLM-DND-CURR-CTRL}	&	0.88	&	{\bf 0.9}	&	{\bf 3.96}	\\
{\color{magenta}LLM-DND-ALL-CTRL}	&	0.87	&	0.88	&  3.92	\\\hline
{\color{brown}GOLD}	&	0.92	&	0.92	&	4.17	\\
    \end{tabular}
\normalsize
    \caption{Average sense, specific and interestingness scores for our systems, and for human-written gold responses}
    \label{tab:manual-eval-overview}
\end{table}

\subsection{Analysis}

\paragraph{Model Comparison.}
Table \ref{tab:manual-eval-overview} shows the average sense, specific and interestingness scores for the systems, and for the human-written gold response. 
All of the \DnD adapted systems outperform the vanilla dialogue system.  
On average, the adapted systems make sense 6.75\% more often than the baseline, are specific 4\% more often, and are 0.37 points more interesting. 
However, the added control features do not seem to differ substantially from the LLM that is adapted to the \DnD data without any control features.  Why then use the control features at all?

\begin{table*}
    \centering
\small
\begin{tabular}{|p{.28\linewidth}|p{.69\linewidth}|}
\hline
{\bf Model}	&	{\bf Output}	\\
\hline
LLM-Dialog	&	moon$\_$monk you can just join our discord conversation. it's free and easier to deal with	\\
\hline
LLM-DND	&	I cast Ice Knife (I'm assuming this is not one of the ones I can't do?)	\\
\hline
LLM-DND-CURRENT-CTRL \newline {\color{red}Responding as Magnus (human fighter, inventory: axe) In combat}	&	I turn from the goblin and go to help Merle and Taako with getting this wagon moving again. Magnus can't see them anymore with such a dense forest. They're likely in a battle right now... I have to find them as fast as I can!	\\
\hline
LLM-DND-CURRENT-CTRL  \newline {\color{red}Responding as Taako (elf wizard, inventory: spellbook) In combat}	&	If the goblins still don't look like they're getting closer, Taako will yell, “Alright you little monsters! Let's see what ya got!” And he'll cast Chill Touch on one of the goblins that isn't dead/on its doorstep. Attack: 22 Damage: 6	\\
\hline
LLM-DND-CURRENT-CTRL  \newline {\color{red}Responding as Mearle (dwarf cleric, inventory: holy symbol) In combat}	&	Merle steps away from the wagon. He draws his sword and takes aim at the hidden goblin! Attack: 27 Damage: 8 He puts his free hand on the pommel of his sword and shouts his battle cry, "For Clanggeddin!" His eyes glow brightly red as he strikes!	\\
\hline
\end{tabular}
\normalsize
\caption{Example model outputs for the next utterance prediction task, given the sample gameplay from Table \ref{gameplay-example} as input. The 3 LLM-DND-CURRENT-CTRL examples use the same model with different control features (in red).}
\label{tab:qualtiative-example-of-next-utterance-prediction}
\end{table*}

\paragraph{Control Features.}
One reason is that the control features are useful to allow the model to role-play.  For example, we can ask the model to play an elf wizard via the control features.
Table \ref{tab:qualtiative-example-of-next-utterance-prediction} gives a qualitative example of this.  The Table shows outputs from two LLMs without control features, plus 3 outputs from our LLM-DND-CURRENT-CTRL model roleplaying different characters.
As input, we gave the models the final seven turns in the game play example from Appendix \ref{example-game-play} –from “With his eagle eyes, Magnus spots two dead horses…” to “... What do you do next?”, and then generate what each model thinks the next utterance ought to be. The vanilla dialogue system with no \DnD data generates an irrelevant reply about Discord. The dialogue system fine-tuned on \DnD data without control feature generates a relevant response.  The dialgoue system that adds conditioning control features allows the model to generate specific replies for each of the different characters.  These replies are relevant to the character classes – the wizard casts a spell, and the Dwarf cleric shouts a battle cry by invoking the name of a Dwarf god.   

\begin{table}
    \centering
\small
    \begin{tabular}{lccc}
{\bf Model}	&	{\bf Sense}	&	{\bf Specific}	&	{\bf Interest}\\\hline
{\color{violet}LLM-Dialog}	&	-0.01	&	-0.01	&	+0.06\\
{\color{orange}LLM-DND}	&	-0.02	&	+0.03	&	+0.4\\
{\color{teal}LLM-DND-PREV-CTRL}	&	+0.02	&	+0.02	&	+0.6\\
{\color{olive}LLM-DND-CURR-CTRL}	&	+0.06	&	{\bf +0.06}	&	{\bf +0.93}\\
{\color{magenta}LLM-DND-ALL-CTRL}	&	{\bf +0.07}	&	{\bf +0.06}	&	+0.81\\\hline
{\color{brown}GOLD}	&	+0.07	&	+0.05	&	+1.02\\
    \end{tabular}
\normalsize
    \caption{Improvements by generating in-character (IC) text rather than out-of-character (OOC) text.  Numbers are IC scores minus OOC scores.  }
    \label{tab:manual-eval-in-character}
\end{table}

\begin{table}
    \centering
\small
    \begin{tabular}{lcc}
\toprule
State variable	&	Majority	&	LLM-DND-GST	\\
\midrule
All	&	.73	&	82	\\
Combat	&	.89	&	.82	\\  
Character Class	&	.58	&	.76	\\
Character Name	&	.58	&	.78	\\
Character Race	&	.75	&	.79	\\
Character Pronouns	&	.58	&	.89	\\
Character Actions	&	.80	&	.85	\\
\bottomrule
    \end{tabular}
\normalsize
    \caption{Average accuracy for our Game State Tracking LLM on the slot-filling our state variables, compared to a majority class baseline.}
    \label{tab:game-state-tracking-accuracy}
\end{table}

\paragraph{In-Character Turns Are More Interesting.}

Among our most impactful control features was the one that allowed systems to generate in-character (IC) versus out-of-character (OOC) turns.  Table \ref{tab:manual-eval-in-character} shows that control models' scores substantially increased on IC turns compared to when their output was generated OOC. The pronounced increase in intersestingness makes sense because IC turns are ones where the players describe their characters in the fictional world often with evocative language, whereas OOC turns usually discuss rules or mechanics.    Our control features allowed the system to intentionally generate IC responses, resulting in substantially improved interestingness scores for those in-character turns.

\section{Game State Tracking Model}
We conducted an experiment to evaluate whether a LLM could be finetuned to perform game state tracking for \DnD using our heuristically annotated game state features.   We trained a new model LLM-DND-GST (Game State Tracking).  It is a LLM-Dialog that has been finetuned on our Play-by-post \DnD gameplay dataset.  As input, it takes all {previous dialog turns and their state variables}, plus {the text of the current turn}, and then it outputs the corresponding {state variables for the current turn}.


We analyzed the accuracy of the LLM-DND-GST model its ability to do slot-filling for each of the individual game states, and compared its performance to a simple baseline that always output the the majority class.  The results are shown in Table \ref{tab:game-state-tracking-accuracy}.
The average accuracy of the dialogue state tracker is better than the majority class baseline, but likely falls short of being useful when it comes to joint accuracy. 
The joint accuracy for LLM-DND-GST is 58\%.  This suggests that accurately tracking the full game state may require additional machinery beyond a finetuned LLM.

\section{Related Work}

Previous work has examined AI to play text adventures games \cite{Haroush2018,Yao2020,Dambekodi}. These games are simpler than  \DnD because they have a limited vocabulary and  more straightforward game states.
Creating text adventure games \cite{Ammanabrolu2020ICCC,Fan2020} is more challenging than playing them, and is similar to the  world-building job of the DM in \DnD.
There has also been work on persona/character generation in stories  \cite{Prabhumoye2019}, and within \DnD~itself \cite{Louis2018}.
Others \cite{LIGHT,Ammanabrolu2020} have realized that NPCs are lacking in their abilities to speak and act in text games.

Findings of the automated story generation community are relevant for \DnD AI systems.
Neural language models have become increasingly more popular for story generation~\cite{Roemmele2018,Martin2018AAAI,Mathewson2019,Hou2019}.
We have also started to see storytelling with transformers~\cite{See2019,Peng2021,Branch2021}.
Transformer-based storytelling systems have even been introduced to the general public thanks to the popularity of AI Dungeon \cite{Walton2019}.
Although neural networks possess a lot of power in terms of what text they generate, they are still limited in their ability to produce longer spans of coherent text.
Many \cite{Fan2018,Yao2019,Ippolito2019,Tambwekar2019,Ammanabrolu2020,Rashkin2020} have improved the coherence of neural storytellers by splitting the generation into two steps: ideation of the story plot, followed by the realization of sentences. 
This {\em controllable story generation} is the focus of a lot of current work in neural automated story generation.

Due to the conversational nature of \DnD, we decided to use a dialog-based system.
Deep neural networks have been used for dialog agents for a while \cite{Serban2016}, with a shift toward using transformers in recent years \cite{Zhang2019,Ghazarian}.
Like in automated story generation and other neural text generation tasks, we are also seeing controllability being an important factor being integrated into systems.
This includes using deep reinforcement learning techniques to guide the dialog toward a goal \cite{Li2016,Saleh2020} or controlling for style \cite{Zhang2018,Smith2020}.


In this paper, we use LaMDA, a transformer-based open-domain dialogue system that builds on the Meena model \cite{meena}.
The original Meena model was an end-to-end model trained on public conversations found on social media.
Controllable text generation with transformers has been seen before with CTRL \cite{Keskar2019}, a language model that is conditioned on a given ``control code'' in addition to the textual history.
This work takes a similar approach.
We 	integrate contextual information such as character descriptions, actions, and in- and out-of-character classifications.


We have finetuned our LaMDA models on data crawled from \DnD~Beyond.
This data contains both in-character and out-of-character dialog and can be used in conjunction with \newcite{Rameshkumar2020}'s dataset from Critical Role (a \DnD~podcast), \newcite{Louis2018}'s dataset from roleplayerguild.com (a \DnD~forum),  \newcite{LIGHT}'s crowdsourced LIGHT dataset, and/or \newcite{storium}'s STORIUM dataset for human+AI collaborative story generation.

\section{Discussion and Conclusions}

We have demonstrated that training on \DnD data results in much higher quality outputs than a vanilla dialogue system (as expected), that controlling the model to generate in-character responses results in substantially more interesting output, and that conditioning on game state information qualitatively results in responses that are appropriate to the character class.  Our preliminary experiments with using the large language models to perform game state tracking show low performance even after finetuning, suggesting that other models may be required for an AI to play \DnD track the full state of the game. 

Although our models are unable to play \DnD fully autonomously by acting as the Dungeon Master,
they could act as an aid for novice DMs.  Since our models can generated evocative, in-character text that is appropriate for the context and the game state, DMs could use it as inspiration  as they narrate the adventure to the other players.  

Here is some model output to inspire your next adventure:  
\begin{quote}
You get a much closer look than the other two... the sarcophagi have the inscriptions of some sort of magic, probably to keep the dead inside, but you can not read them to save your life.
\end{quote}
What will you do next?  Download our dataset\footnote{\url{https://www.cis.upenn.edu/~ccb/dnd-data.html}} to start your new adventure!

\section{Limitations}

One limitation of our human evaluation is that it is a static evaluation.  The raters are simply reading the outputs of the model, and there is no interactive evaluation wherein they engage in gameplay with the system.  An interactive user-study would be required before any claims could be made about how well AI is able to play \DnD alongside human players. 

Because our state information was created heuristically, it therefore potentially contains errors.  It is also incomplete. There are several kinds of state tracking variables that would be useful to include, but were not possible to heuristically extract from our data. %
To address this problem in the future, we have begun a collaboration with the developer of Avrae,
which is a Discord bot for playing \DnD online.  Avrae contains many state variables that are missing from our current annotations, such as HIT points and slot-filler values for attacks.

\section{Acknowledgments}

We would like to thank Antony Pegg of Fandom for granting us permission to use \DnD Beyond's forum data for research purposes. 

Thank you to Rich Burlew and forum moderator truemane for granting us permission to crawl the Giant in the Playground forum and to build models using the forum posts. 

We would like to thank the many Google Researchers who gave valuable input on this project, especially Dipanjan Das and Suma Bailis.  

Chris Callison-Burch is grateful for having the opportunity to be a part-time visiting researcher at Google for two years.  It was amazing to spend time among such incredibly smart people, and it was eye opening to see large LMs before they became widely available. Thanks! 


\bibliographystyle{acl_natbib}
\bibliography{main}

\appendix

\begin{table*}
\small
\begin{tabular}{p{.15\linewidth}p{.38\linewidth}p{.38\linewidth}}
\toprule
{\bf Player\newline (character)}	&	{\bf Game Dialogue}	&	 {\bf \DnD Game Description and\newline AI challenges}\\
\hline
Griffin (DM)	&	A dwarf named Gundren Rockseeker has hired you to transport a wagonload of provisions to the rough-and-tumble settlement of Phandalin, which is a couple days' travel to the southeast. A day and a half after leaving, you turn off the high road that connects the major cities on the coast onto a smaller trail that will lead you to Phandalin. This trail is not as well maintained, and bandits and outlaws have been known to lurk along the trail.	&	{\it This game is based on the  \DnD starter adventure called ``Lost Mine of Phadelver”.  The adventure book is a mixture of rules and “boxed text” which is descriptive text for the DM to read aloud or paraphrase. See the appendix for the text that the DM is consulting.} AI challenges: {\bf Generation of stories and descriptive text}\\
Griffin (DM)	&	Roll a perception check for me.  Perception is a wisdom skill, so be sure to add your wisdom modifier.	& {\it	The previous text was descriptive text.  Here the DM is asking the players to perform a game mechanic and referencing a game rule. This is called “out of character” dialogue. } AI challenges: {\bf Knowledge base population (extraction of rules from a rulebook)}	\\
Clint (out of character)	&	I got an eight.	&	{\it Clint has rolled his dice. The number is low so his character fails the check.} AI challenges:  { \bf Multi-party dialogue}\\
Justin (out of character)	&	I got a six.	&	{\it Justin also fails. Neither character sees the thing that requires the perception  check.}	\\
Travis (out of character)	&	I rolled a natural twenty plus my wisdom modifier is 23.	&	{\it Travis rolls high number and succeeds on the check.} AI challenges: {\bf Understanding rules, determining success or failure} \\
Griffin (DM)	&	With his eagle eyes, Magnus spots two dead horses lying in the middle of the road about 200 feet ahead of you.	&	{\it The DM describes what happens as a result of the success.  AI challenges:} {\bf Reasoning about consequences of success or failure, descriptive text generation} \\
Travis \newline(in-character as Magnus)	&	I stop the wagon and motion silently to get the attention of Merle and Taako, and kinda pull them up towards the front of the wagon.	&	{\it Travis is describing what he is doing using “in character” language.} AI challenges: {\bf Persona-based chat} \\
Griffin (DM)	&	As you warn them that shit has gone south, you notice a few goblins crouching in a part of the shaded woods off to the side of the road.  Two of the goblins begin charging your wagon.	&	{\it The DM describes the start of a battle with several monsters.} AI challenges: {\bf State tracking (in combat v. out of combat).} \\
Travis (out of character)	&	How many goblins are there?	&	AI challenges: {\bf Question answering, state tracking (how many monsters).}	\\
Griffin (DM)	&	There are three goblins; two of them are rushing the group, one is pretty heavily obscured by the brush, probably about 40 feet out, sort of between you and the dead horses laying in the middle of the road.	&	AI challenges: {\bf Question answering, Descriptive text generation from game state.}	\\
Clint (Merle)	&	I will cast sacred flame at the nearest one.  If it fails a dexterity saving throw, it takes 6 points of damage.	&	{\it Clint chooses an action based on what is allowed for his character class. He describes the rule that governs the spell in an out-of-character fashion.} AI challenges: {\bf Intent detection (perform attack action against a particular goblin)} \\
Griffin (DM)	&	You attack. You launch some fire onto the goblin closest to the wagon. And with that, he looks like he is on death's door. And the other goblin that you can see, the one that‘s not in the brush somewhere, just sort of stops in his tracks.  What do you do next?	&	{\it The DM rolls for the monster, updates the state of its health meter, and describes the result of Merle's attack.} AI challenges: {\bf Reasoning about rules, state tracking (monster's HIT points), descriptive text generation.} \\
\bottomrule
\end{tabular}
\normalsize
\caption{Example dialogue from a game of \DnD with explanations of what is happening and comments on potential challenges for AI}\label{gameplay-example}
\end{table*}

\section{Example \DnD Game Session}\label{example-game-play}

Instead of the game being a series of moves on a game board, RPGs \DnD are language-based.  Players create characters that have a class (wizard, fighter, thief, etc.) that denotes their abilities, and a fantasy race (elf, dwarf, human, etc.).  Players describe what they want their character to do and roll dice to determine if they are successful. The dungeon master (DM) acts as the narrator who shapes the overall story.  The DM describes scenarios and locations, and takes on the role of non-player characters (NPCs), and monsters.  

A common element to the game play is an encounter with monsters.  Battles are governed by rules, and unfold in a turn-based fashion where the DM controls the monsters and each player controls their character. 
Each player and monster has a health meter (called their HIT points), an armor class (which indicates the threshold of the dice roll needed to damage them), and a set of possible attack or move actions.  

Table \ref{gameplay-example} provides example dialogue from a game of \DnD being played between 3 players – Travis (playing a human fighter named Magnus Burnsides), Clint (playing Merle Highchurch, a dwarf cleric), Justin (playing Taako an elf wizard), and DM Griffin.
We add comments about each dialogue turn to describe what is happening in the game, and to highlight the challenges that would need to be addressed if an AI system were to play the game either as a player or as the DM. 

The game session is taken from the podcast {\it The Adventure Zone}.  In this episode, the hosts are playing an adventure module called {\it  Lost Mine of Phadelver}, an expert of which is given in Appendix \ref{lost-mine-exerpt}. In the first episode of the podcast\footnote{\url{https://maximumfun.org/episodes/adventure-zone/ep-1-here-there-be-gerblins-chapter-one/}}, the hosts explain the rules of \DnD.

\section{Lost Mine of Phadelver Adventure}\label{lost-mine-exerpt}

Here is an excerpt from the adventure book that the Dungeon Master was using in our example game play.  The adventure book provides boxed text, which is descriptive text to be read aloud verbatim or to paraphrase.  It also gives details about the combat that is about to ensure,  and links to relevant game rules (like stealth checks, and statistics about the monsters that the characters will be in combat with). 

\small
    
The adventure begins as the player characters are escorting a wagon full of provisions and supplies from Neverwinter to Phandalin. The journey takes them south along the High Road to the Triboar Trail, which heads east (as shown on the overland map). When they're a half-day's march from Phandalin, they run into trouble with goblin raiders from the Cragmaw tribe.

Read the boxed text when you're ready to start. If you create a different adventure hook, skip to the second paragraph and adjust the details as necessary, ignoring the information about driving the wagon.

\begin{quote}
In the city of Neverwinter, a dwarf named Gundren Rockseeker asked you to bring a wagonload of provisions to the rough-and-tumble settlement of Phandalin, a couple of days' travel southeast of the city. Gundren was clearly excited and more than a little secretive about his reasons for the trip, saying only that he and his brothers had found “something big,” and that he'd pay you ten gold pieces each for escorting his supplies safely to Barthen's Provisions, a trading post in Phandalin. He then set out ahead of you on horse, along with a warrior escort named Sildar Hallwinter, claiming he needed to arrive early to “take care of business.”

You've spent the last few days following the High Road south from Neverwinter, and you've just recently veered east along the Triboar Trail. You've encountered no trouble so far, but this territory can be dangerous. Bandits and outlaws have been known to lurk along the trail.

You've been on the Triboar Trail for about half a day. As you come around a bend, you spot two dead horses sprawled about fifty feet ahead of you, blocking the path. Each has several black-feathered arrows sticking out of it. The woods press close to the trail here, with a steep embankment and dense thickets on either side.

\end{quote}

Four goblins are hiding in the woods, two on each side of the road. They wait until someone approaches the bodies and then attack.

This will likely be the first of many combat encounters in the adventure. Here are the steps you should follow to run it effectively:

\begin{itemize}
\item Review the goblin stat block. Since the goblins are hiding, you'll need their Stealth skill modifier: +6.
\item Check to see who, if anyone, is surprised. The party can't surprise the goblins, but the goblins might surprise some or all of the characters. Make a Dexterity (Stealth) check for the goblins: roll one d20 for all of them, add their Stealth skill modifier (+6) to the roll, and compare the total to the characters' passive Wisdom (Perception) scores. A character whose score is lower than the goblins' check total is surprised and therefore can't do anything on his or her first turn in the combat (see “Surprise” in the Basic Rules).
\item Use the initiative rules in the Basic Rules to determine who acts first, second, third, and so on. Keep track of everyone's initiative count on a piece of paper.
\item When the time comes for the goblins to act, two of them rush forward and make melee attacks while two goblins stand 30 feet away from the party and make ranged attacks. The goblins' stat block contains the information you need to resolve these attacks. For more information on what the goblins can do on their turn, see chapter 9, “Combat,” in the Basic Rules.
When three goblins are defeated, the last goblin attempts to flee, heading for the goblin trail
\end{itemize}

\normalsize

\begin{table*}
    \centering
    \begin{tabular}{ccccccc}
State variable	&	Model	&	Type	&	Multi-valued	&	Availability	&	Evaluation metric	&	Performance	\\ \hline
Character	&	Span labeller	&	Text	&	No	&	42\%	&	-	&	-	\\
Race	&	Classifier	&	Text	&	No	&	$>$58\%	&	Macro AUC	&	0.45	\\
Class	&	Classifier	&	Text	&	No	&	$>$75\%	&	Macro AUC	&	0.71	\\
Pronous	&	Classifier	&	Text	&	No	&	42\%	&	Macro AUC	&	0.92	\\
Inventory	&	Span labeller	&	Text	&	Yes	&	11\%	&	-	&	-	\\
In combat?	&	Classifier	&	Score	&	No	&	100\%	&	Accuracy	&	0.91	\\
Action	&	Classifier	&	Text	&	Yes	&	20\%	&	Macro AUC	&	0.92	\\
    \end{tabular}
    \caption{The estimated performance of our CNN classifier on predicting state values for turns where our rule-based heuristics did not predict a value}
    \label{tab:cnn-performance-predicting-state-variables}
\end{table*}

\section{Estimated accuracy of predicted state variables}\label{cnn-state-values}

In addition to the heuristics that we used to recover state variables for each turn in the game (described in Section \ref{heuristic-annotation}), we used a CNN to fill in state values when our heuristics did not fire. 
Table \ref{tab:cnn-performance-predicting-state-variables} estimates gives an estimate of the CNN's performance on filling in the state variables where the rule-based heuristic did not extract a value. 
The CNN classifier only uses current post text as input (no additional context).

\section{Annotation Guidelines and Annotation Interface}\label{annotation-details}

\subsection{Annotation task}
In this task, you will see part of a conversation between a few people playing \DnD. The players and their characters are listed at the beginning of the conversation.  The conversations that are shown as context are real conversations from players. Your job is to read the context and then rate different responses for a player/character given conversational context. Please note that the context you are given represents only a part of the players' past conversations/interactions with one another during the game.

For each response, you would be asked the following questions.
\begin{itemize}
    \item Does the response make sense?
    \begin{itemize}
        \item Use your common sense here. Is the response completely reasonable in terms of the rules of \DnD?
        \item The response “makes sense” if it is cohesive as a standalone statement, consistent with the rules of the game, and the elements/entities mentioned are plausible, given the prior context.
        \item If anything seems off—not fluent, confusing, illogical, out of context, or wrong according to the rules of \DnD —then rate it as Does not make sense.
If in doubt, choose Does not make sense.
    \end{itemize}
    \item Is the response specific?
        \begin{itemize}
        \item You may be asked to assess whether the response is specific to a given context. In other words, do you think that the response represents a good thing for the character to do now?
        \item The response is "specific" if it flows logically from the narrative established by the prior context.
            \begin{itemize}
            \item Note: It is possible for a response to "make sense" (due to being cohesive, consistent and plausible in and of itself), but be marked "not specific" when it is not a logical next step in the overall game progression.
            \item Note: "Specific" for the purposes of this task does not have to do with how detailed the response is per se; a response can be fairly general in its language, but still qualify as "specific" when it is a logical next step in the overall game progression.
            \end{itemize}
    \end{itemize}
    \item How interesting is the response?\begin{itemize}
        \item You may be asked to score the response for its interestingness on a scale of 10.

Choose a high score for “Interesting” if the response would likely catch someone's attention or arouse curiosity in the game; or it is insightful, creative, or witty with respect to the game. If the response is monotonous and predictable, or if you're unsure, then it is Less Interesting.
            \end{itemize}
\end{itemize}

\subsection{Annotation Interface}
\begin{figure*}
    \centering
    \includegraphics[width=\linewidth]{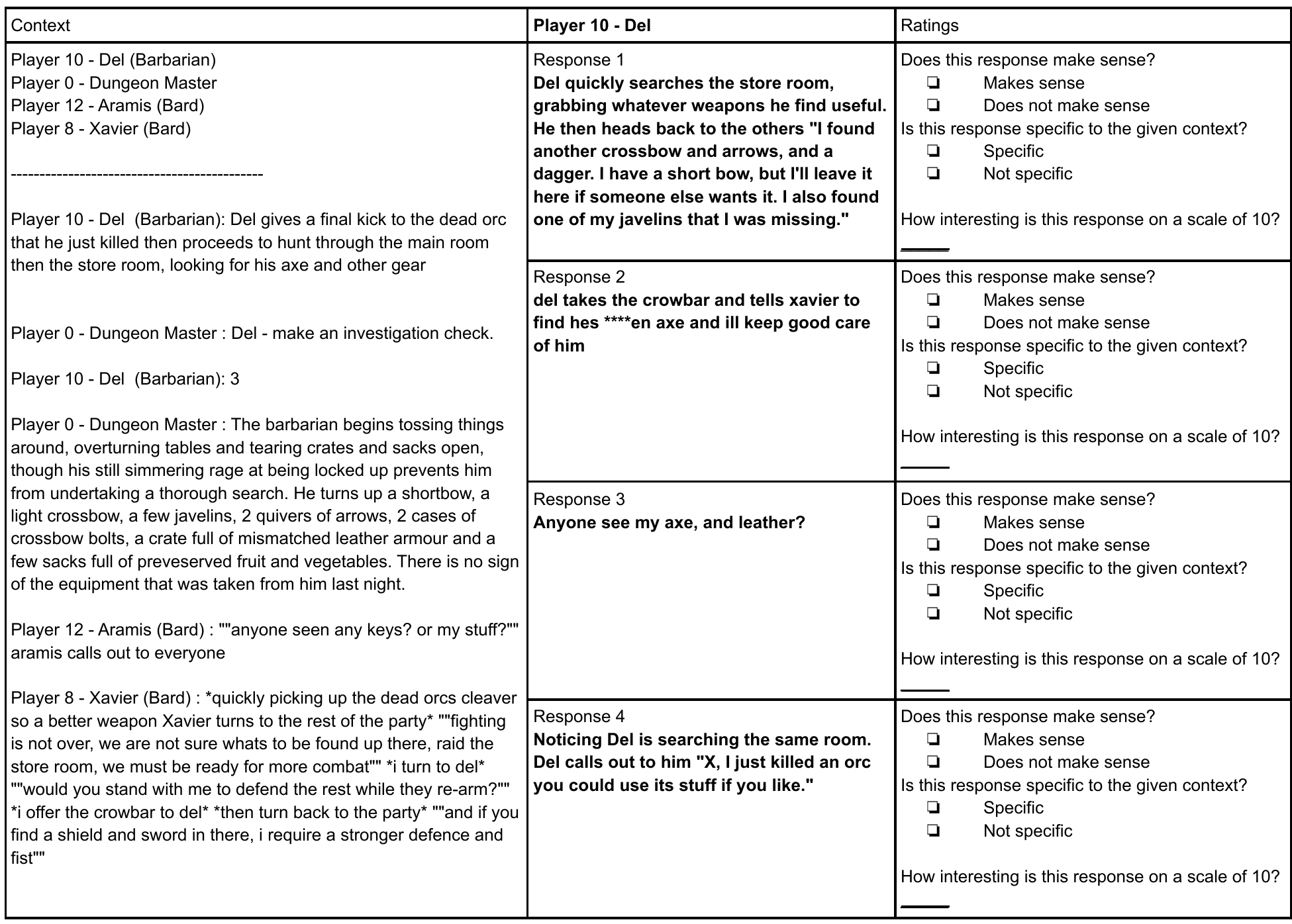}
    \caption{The user interface that our raters used to evaluate the quality of our model's next utterance prediction. }
    \label{fig:annotation-interface}
\end{figure*}

A mock up of the annotation user interface is given in Figure \ref{fig:annotation-interface}.  

\subsection{Survey of Raters}

We recruited raters who had a background in role playing games and an understanding of the fantasy genre.  We surveyed our raters, asking them the following questions:

\begin{enumerate}
\item  Have you ever played Dungeons and Dragons or another role playing game before?
\item If so,
    \begin{itemize}
	\item roughly how many times have you played
	\item were you a player or a game master or both
    \end{itemize}
\item If not,
    \begin{itemize}
	\item  what kind of exposure do you have to Dungeons and Dragons? (For example, have you seen it referred to in TV or movies)
    \end{itemize}

\item Are you a fan of the fantasy genre (like Lord of the Rings)?

\end{enumerate}

Our 6 raters responded to the survey as follows:
5 out of the 6 have played \DnD or another role playing game before.
All 5 of those who have played \DnD/other role playing games before have played more than 6 times.
Of the 5 who have played \DnD/other role playing games before, 3 played as both Game Master and Player.
For the one who had not played \DnD/other role playing games, they indicated they had not had much exposure to \DnD through TV or other channels.
All 6 answered that they were fans of the fantasy genre.

\end{document}